\documentclass{article} 
\usepackage{iclr2027_conference,times}

\usepackage{amsmath,amsfonts,bm}

\def\eqref#1{equation~\ref{#1}}

\def\1{\bm{1}}

\DeclareMathAlphabet{\mathsfit}{\encodingdefault}{\sfdefault}{m}{sl}
\SetMathAlphabet{\mathsfit}{bold}{\encodingdefault}{\sfdefault}{bx}{n}

\usepackage{hyperref}
\usepackage{url}
\usepackage[pdftex]{graphicx}
\usepackage{booktabs}
\usepackage{multirow}
\usepackage{subcaption}
\usepackage{xcolor}
\usepackage{algorithm}
\usepackage{algpseudocode}
\usepackage{tabularx}
\usepackage{caption}
\usepackage{marvosym} 
\usepackage{fontawesome5}
\usepackage{simpleicons}

\newsavebox{\ablationbox}
\newlength{\ablationheight}

\algrenewcommand\algorithmiccomment[1]{\hfill\small\textcolor{gray}{\textit{// #1}}}

\definecolor{mycolor_green}{HTML}{E6F8E0}

\title{OmniOPD: Stable On-Policy Distillation for Joint Audio-Video Generation}

\author{Zihao Fan$^{*}$, Xin Lu$^{*}$, Jie Huang\textsuperscript{\ddag}, Hexin Zhang, Xueyang Fu\textsuperscript{\Letter}, Zheng-Jun Zha
\\
University of Science and Technology of China\\
\raisebox{-0.4ex}{\includegraphics[height=1em]{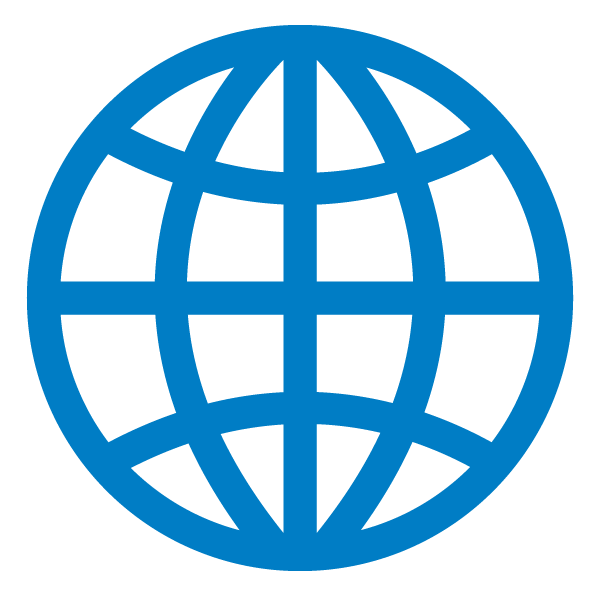}}\hspace{1em}\href{https://fanzh03.github.io/OmniOPD}{\texttt{Project Page}} 
\vspace{0.2cm} 
\hspace{1em}
\raisebox{-0.4ex}{\includegraphics[height=1em]{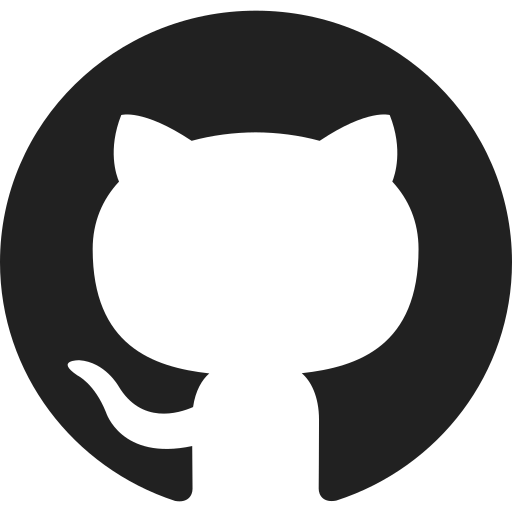}}\hspace{1em}\href{https://github.com/fanzh03/OmniOPD}{\texttt{Code}} 
\hspace{0.2cm}
\vspace{0.2cm} 
\raisebox{-0.4ex}{\includegraphics[height=1em]{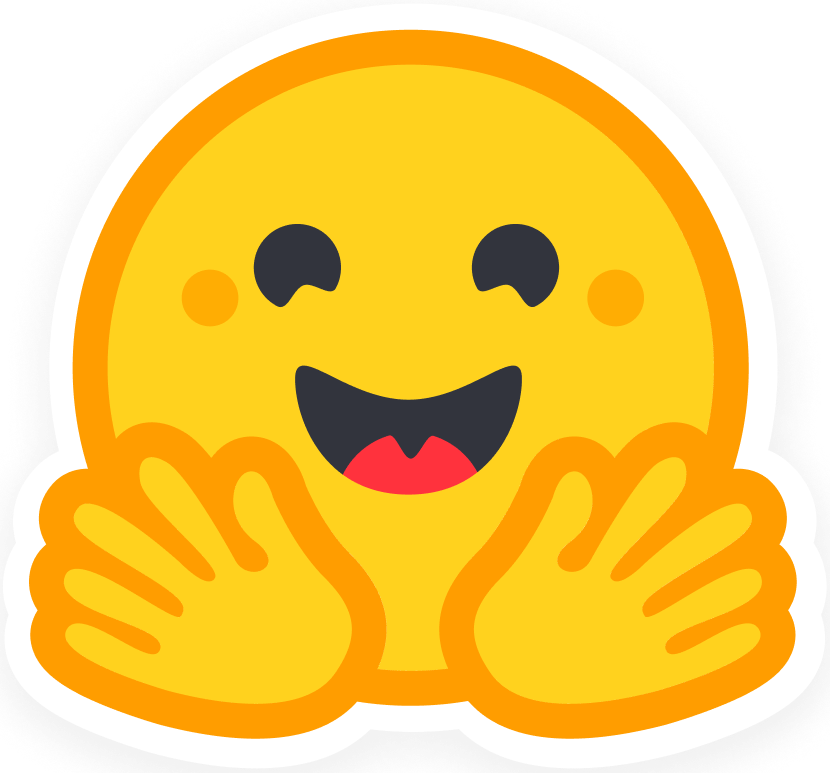}}\hspace{1em}\href{https://huggingface.co/fanzh03/OmniOPD}{\texttt{Model}}
}

\iclrfinalcopy 
\begin{document}
\maketitle
\lhead{}\renewcommand{\headrulewidth}{0pt}
{\renewcommand{\thefootnote}{}\footnotetext{\textsuperscript{\ddag}Project leader,\textsuperscript{\Letter}Corresponding author.}
}

\maketitle
\begin{abstract}
Reinforcement learning (RL) elevates joint audio-video generation, but optimizing heterogeneous, modality-specific audio and video rewards often introduces cross-modal interference. On-policy distillation (OPD) alleviates this issue by transferring the capabilities of specialized audio and video RL teachers into a single student model. 
However, a significant imbalance in cross-modal token counts leaves audio with sparse supervision, causing unstable distillation. 
We observe that naively increasing the audio loss weight fails to solve this problem since audio learning relies heavily on visual context, indicating richer visual cues are crucial to offset sparse audio signals. 
To address this, we propose OmniOPD, a modality-wise on-policy distillation framework designed for balanced and stable audio-video generation. 
First, we construct privileged visual information using asymmetric noise scheduling, allowing the audio teacher to generate more informative targets from cleaner video context. 
Second, we introduce grouped multi-noise sampling to expand audio supervision coverage and stabilize training. 
Third, we introduce timestep-wise importance weighting based on teacher-student transition divergence, preserving reliable supervision while down-weighting unreliable learning signals, to stabilize fluctuations in audio learning.
Evaluated on JavisBench and VBench demonstrate OmniOPD preserves audio quality while significantly enhancing video generation, achieving effective joint distillation.
\end{abstract}

\section{Introduction}
\label{sec:intro}

Joint audio-video generation has made substantial progress through unified models and cross-attention mechanisms~\citep{liu2025javisdit}. Reinforcement learning (RL) provides an effective post-training approach for further improving audio or video quality by optimizing a single reward signal for a specific task~\citep{zheng2025diffusionnft,liu2025flowgrpo}. However, jointly optimizing multiple modalities and rewards introduces conflicting objectives and imbalanced gradients across modalities, leading to cross-modal interference and hindering improvements in overall performance~\cite{GDPO,zhang2026omninft}.

\begin{figure}
    \centering
    \includegraphics[width=1.0\linewidth]{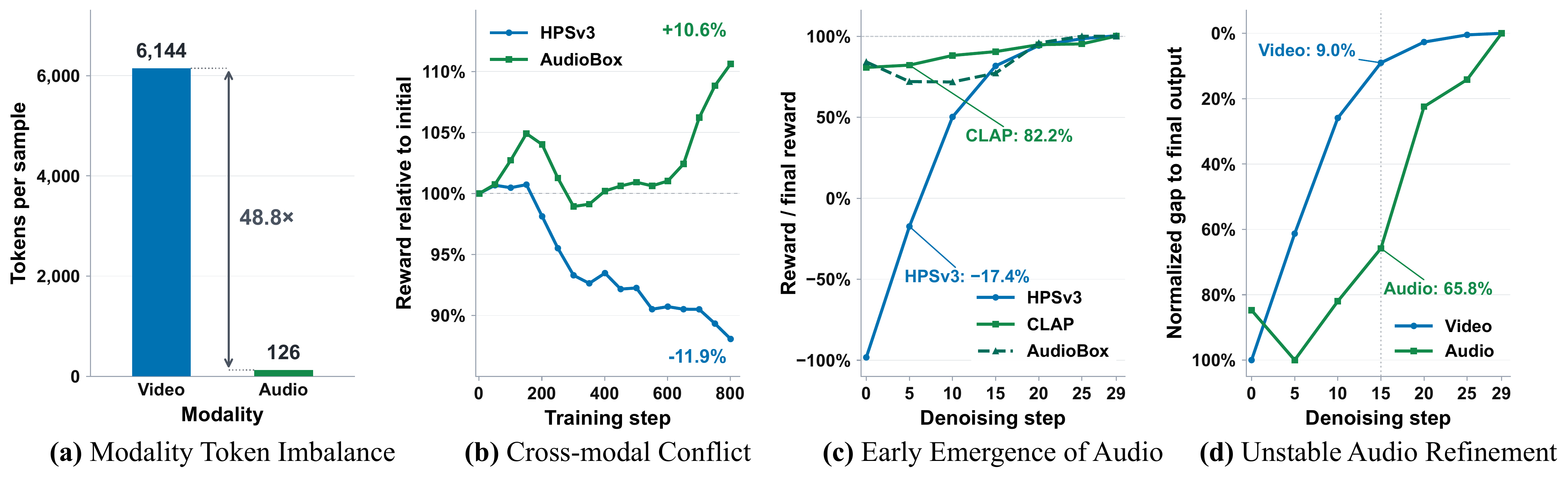}
    \caption{\textbf{Motivation analysis of audio-video joint learning.} (a) Audio and video exhibit a substantial token imbalance, leading to training instability. (b) Optimization guided solely by audio rewards is unstable and can degrade video quality. (c)(d) Audio performance improves early during denoising, but audio refinement remains less stable than video refinement.}
    \label{fig:prior}
\end{figure}

On-policy distillation (OPD)~\citep{thinkingmachines2026opd} offers an alternative by decoupling specialist optimization from joint capability integration~\cite{diffusionopd}. Modality-specialized teachers are first optimized independently using their respective rewards, and their capabilities are then transferred to a unified student through dense supervision along the student’s own generation trajectories. However, strong teachers alone do not guarantee balanced capability transfer. In joint audio-video generation, audio typically occupies substantially fewer tokens than video, leaving it with comparatively sparse supervision, as illustrated in Fig.~\ref{fig:prior}(a). This imbalance can destabilize audio distillation, making the coordination of heterogeneous supervision a central challenge for effective joint learning.

We conduct a comprehensive analysis of the coupling between audio and video in joint learning (see Sec.~\ref{sec:motivation} for details).
First, audio learning exhibits substantial fluctuations under comparatively sparse supervision, and audio-only RL leads to unstable training dynamics (Fig.~\ref{fig:prior}).
Second, the instability during audio learning also degrades video quality (Fig.~\ref{fig:prior}(b)), revealing the coupling between the two modalities within a shared model. 
Third, optimizing video quality before audio leads to more stable audio learning than optimizing audio alone, as illustrated in Fig.~\ref{fig:reward}. 
This suggests that stronger video capability supports audio learning and motivates using visual information to strengthen audio supervision. 
Together, these observations point to the need for more informative and reliable audio supervision to support balanced joint learning.

Motivated by these observations, we propose OmniOPD, a modality-wise on-policy distillation framework that improves the stability and balance of audio-video learning (see Fig,~\ref{fig:pipeline}). 
First, privileged visual guidance constructs asymmetric noise conditions, allowing the audio teacher to produce more informative targets from cleaner video context while sharing the same audio input with the student. 
Second, grouped multi-noise sampling collects diverse and complementary audio supervision, broadening coverage of the audio state distribution and stabilizing audio distillation.
Third, timestep-wise importance weighting uses local teacher–student transition divergence to assess supervision reliability, retaining full supervision within a trust region and smoothly down-weighting high-mismatch supervision beyond it.
The resulting audio objective is jointly optimized with video distillation to integrate modality-specialized capabilities into a unified student.

Our contributions are summarized as follows:
\begin{itemize}
    \item We comprehensively analyze audio-video coupling phenomena, identifying sparse and unstable audio supervision as a key challenge in joint audio-video distillation and highlighting the role of visual context in supporting audio learning.
    \item We propose OmniOPD, which combines privileged visual guidance, grouped multi-noise sampling, and importance weighting to stabilize joint audio-video on-policy distillation.
    \item Extensive experiments demonstrate that OmniOPD stabilizes joint distillation and achieves balanced gains in audio quality, video quality, and audio-video synchronization.
\end{itemize}

\section{Related Work}
\label{sec:related}

\subsection{Joint Audio-Video Generation}

Joint audio-video generation~\citep{liu2026javisditpp,low2025ovi,chefer2026selfflow} models the two modalities together to produce synchronized audiovisual content. Early work, MM-Diffusion~\citep{ruan2022mmdiffusion}, couples audio and video denoisers through cross-modal attention. Subsequent approaches advance joint synthesis through diffusion transformers and the integration of pretrained modality experts. JavisDiT~\citep{liu2025javisdit} introduces hierarchical spatio-temporal priors to improve synchronization, while UniVerse-1~\citep{wang2025universe} combines pretrained audio and video experts through block-level stitching. LTX-2~\citep{hacohen2026ltx2} adopts an asymmetric dual-stream transformer with bidirectional cross-attention, supporting modality-specific generation and cross-modal interaction within a unified model. These methods establish joint synthesis architectures and cross-modal communication mechanisms. Our work studies how a joint model can absorb improvements from modality-specialized teachers through stable distillation.

\subsection{Reinforcement Learning for Generative Models}

Reinforcement learning provides a post-training approach for aligning generative models with task-specific rewards and human preferences.
Early methods such as DDPO~\citep{black2023ddpo}, DPOK~\citep{fan2023dpok}, and ReFL~\citep{xu2023imagereward} improve diffusion models through reward-driven fine-tuning. Diffusion-DPO~\citep{wallace2023diffusion} further enables alignment through direct optimization on pairwise preference data. GRPO~\citep{GRPO} has subsequently been adapted to visual generation: Flow-GRPO~\citep{liu2025flowgrpo} and DanceGRPO~\citep{xue2025dancegrpo} employ stochastic generation trajectories and group-relative rewards to optimize flow-based models. For multi-reward optimization, GDPO~\citep{GDPO} normalizes individual rewards separately to preserve their distinct contributions and improve training stability. DiffusionNFT~\citep{zheng2025diffusionnft} introduces a forward-process formulation that incorporates reward feedback through an implicit policy-improvement objective. Building on this formulation, OmniNFT~\citep{zhang2026omninft} extends this framework to joint audio-video generation with modality-aware optimization.

\begin{figure}[!t]
    \centering
    \includegraphics[width=1.0\linewidth]{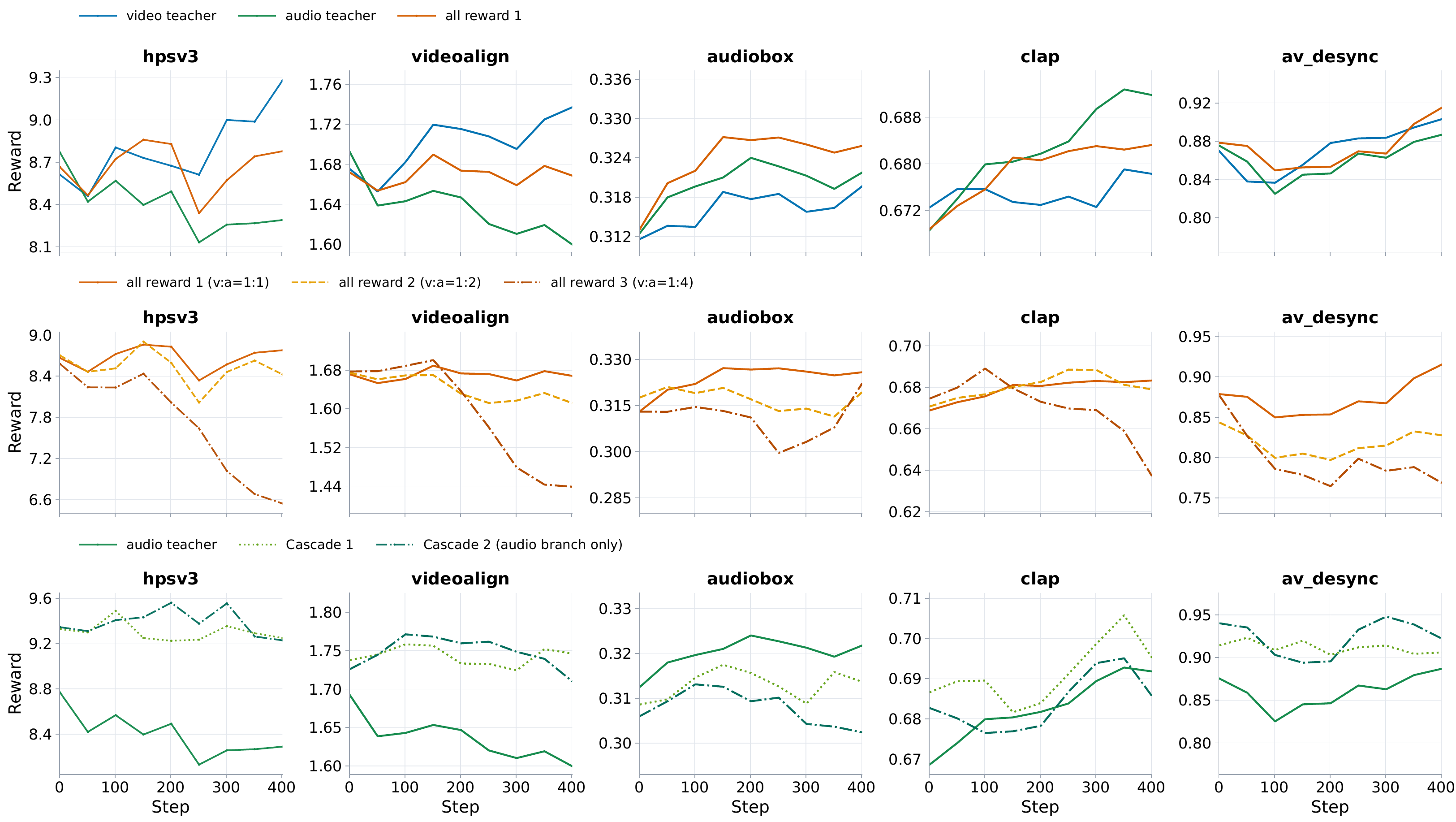}
    \caption{\textbf{A comprehensive analysis of audio-video coupling during optimization.} The top row compares optimization with video-only, audio-only, and joint rewards. The middle row compares three joint-reward configurations, including increased audio reward weights. The bottom row compares audio-only optimization with two Cascade variants that optimize video first and then audio. (1) Audio-only reward optimization is unstable and can degrade video quality; increasing audio reward weights fails to resolve these conflicts and can further destabilize training.(2) Joint-reward optimization yields limited gains due to conflicts among reward objectives across modalities. (3) Sequential video-to-audio optimization stabilizes training, suggesting that cleaner and more stable video context is important for audio learning.}
    \label{fig:reward}
\end{figure}

\subsection{On-Policy Distillation}
On-policy distillation~\citep{thinkingmachines2026opd} learns from teacher feedback on student-generated states, with language-model studies examining divergence objectives and sample reuse~\citep{agarwal2023gkd,gu2023minillm,ko2024distillm}. MOPD~\citep{ma2026mopd} integrates independently trained specialists, while Open-MOPD~\citep{gao2026openmopd} studies imbalanced supervision budgets. TrOPD~\citep{xing2026tropd} addresses unreliable token supervision under distribution mismatch. For continuous generation, Flow-OPD~\citep{flowopd} consolidates specialist teachers along student rollouts, and DiffusionOPD~\citep{diffusionopd} derives closed-form stochastic transition losses and deterministic mean-matching surrogates. Recent extensions address heterogeneous teachers, output and representation targets, and reward-guided self-distillation~\citep{wei2026stepopd,zhou2026diffusionopsd}. OmniOPD builds on vanilla OPD to stabilize joint audio-video distillation.

\section{Motivation}
\label{sec:motivation}
We independently train a video teacher and an audio teacher using only video-related and audio-related rewards, respectively, while the All reward variants are trained with a mixture of both.

\paragraph{Audio instability affects both modalities.}
As shown in Fig.~\ref{fig:reward}(top), audio-only reward optimization exhibits unstable training dynamics, with deteriorating audio quality accompanied by declining video quality even when optimization targets audio. This observation reveals the coupling between the two modalities within a shared model and highlights the importance of stabilizing audio learning for joint generation.

\paragraph{Increasing audio reward weights does not resolve the imbalance.}
As shown in Fig.~\ref{fig:reward}(middle), joint-reward optimization yields limited gains, and increasing audio reward weights can further degrade both modalities. Simply upweighting audio rewards does not address the underlying imbalance in supervision; instead, it amplifies the influence of unstable audio learning signals, potentially undermining joint optimization.

\paragraph{Video-first optimization stabilizes audio learning.}
As shown in Fig.~\ref{fig:reward}(bottom), optimizing video first and subsequently training audio substantially alleviates the instability observed in audio-only optimization. This finding suggests that stronger video capability provides a more favorable context for audio learning, motivating the use of cleaner and more stable visual information to strengthen audio supervision.

Together, these findings highlight sparse and unstable audio supervision as a key bottleneck in joint audio-video learning and suggest that cleaner visual context can help alleviate it. They motivate OmniOPD's three complementary components: privileged visual guidance to provide more informative audio targets, grouped sampling to broaden supervision coverage, and timestep-wise importance weighting to reduce the influence of unreliable learning signals.

\section{Preliminaries}
\label{sec:method_prelim}

\subsection{Joint Audio-Video Flow Matching}
\label{sec:joint_avgen}
We consider a joint audio-video flow matching~\citep{lipman2023flow} model with audio and video latents $x_0^a$ and $x_0^v$. Following standard joint audio-video flow matching~\citep{hacohen2026ltx2}, each modality is independently perturbed by a Gaussian endpoint $x^m_{1}\sim\mathcal{N}(0,I)$, $m\in\{a,v\}$, while sharing the same flow timestep $t\in[0,1]$:
\begin{equation}
x_{t}^m=(1-t)x_{0}^m+t x_{1}^m,\qquad m\in\{a,v\}.
\end{equation}

The audio and video velocity fields are jointly predicted as $v_\theta^m(x_t^a,x_t^v,t,c)$, where each modality is conditioned on the complete audio-video state. The joint flow-matching objective is
\begin{equation}
\mathcal{L}_{\mathrm{FM}}
=
\mathbb{E}
\left[
\sum_{m\in\{a,v\}}
\left\|
v_m(x_t^a,x_t^v,t,c;\theta)
-
(x_1^m-x_0^m)
\right\|_2^2
\right].
\label{eq:joint_fm}
\end{equation}

\subsection{On-Policy Distillation}
\label{sec:opd}

On-policy distillation~\citep{thinkingmachines2026opd} lets the student autoregressively generate a full sequence from its own policy, and then trains the student to match the teacher on the prefixes that the student itself visits.
Concretely, let $\tau=(x_{t_0},\ldots,x_{t_N})\sim P_\theta$ be a state trajectory generated by the student with parameters $\theta$. Let $p_S(\cdot\mid x_{t_k})$ and $p_T(\cdot\mid x_{t_k})$ denote the student and frozen teacher one-step transition kernels, respectively. Following DiffusionOPD~\citep{diffusionopd}, we match these kernels along student trajectories on $1=t_0>\cdots>t_N=0$. With $\mathcal T=\{0,\ldots,N-1\}$, the objective is
\begin{equation}
\mathcal L_{\mathrm{OPD}}(\theta)
=\mathbb E_{\tau\sim P_\theta}
\left[\sum_{k\in\mathcal T}
D_{\mathrm{KL}}(p_S(\cdot\mid x_{t_k})\|p_T(\cdot\mid x_{t_k}))\right].
\label{eq:opd_loss}
\end{equation}
For per-step Gaussian transitions $p_M(\cdot\mid x_{t_k})=\mathcal N(\mu_{M}(x_{t_k}),\bar{\sigma}_k^2 I)$, $M\in\{S,T\}$, the means $\mu_{M}(x_{t_k})$ differ while the variance $\bar{\sigma}_k^2$ is shared. The local Kullback-Leibler divergence is
\begin{equation}
D_{\mathrm{KL}}(p_S(\cdot\mid x_{t_k})\|p_T(\cdot\mid x_{t_k}))
=\frac{\|\mu_{S}(x_{t_k};\theta)-\mu_{T}(x_{t_k})\|_2^2}{2\bar{\sigma}_k^2}.
\label{eq:local_kl_def}
\end{equation}

For the stochastic flow transitions in Flow-OPD~\citep{flowopd}, let $w(t)$ represents the time-adaptive scaling factor. The local divergence becomes
\begin{equation}
D_{\mathrm{KL}}
= w(t)\left\|v_S(x_{t_k};\theta)-v_T(x_{t_k})\right\|_2^2,
\label{eq:flow_local_kl}
\end{equation}
where $v_S(x_{t_k};\theta)$ and $v_T(x_{t_k})$ are the student and teacher velocity predictions.

\begin{figure}
    \centering
    \includegraphics[width=1.0\linewidth]{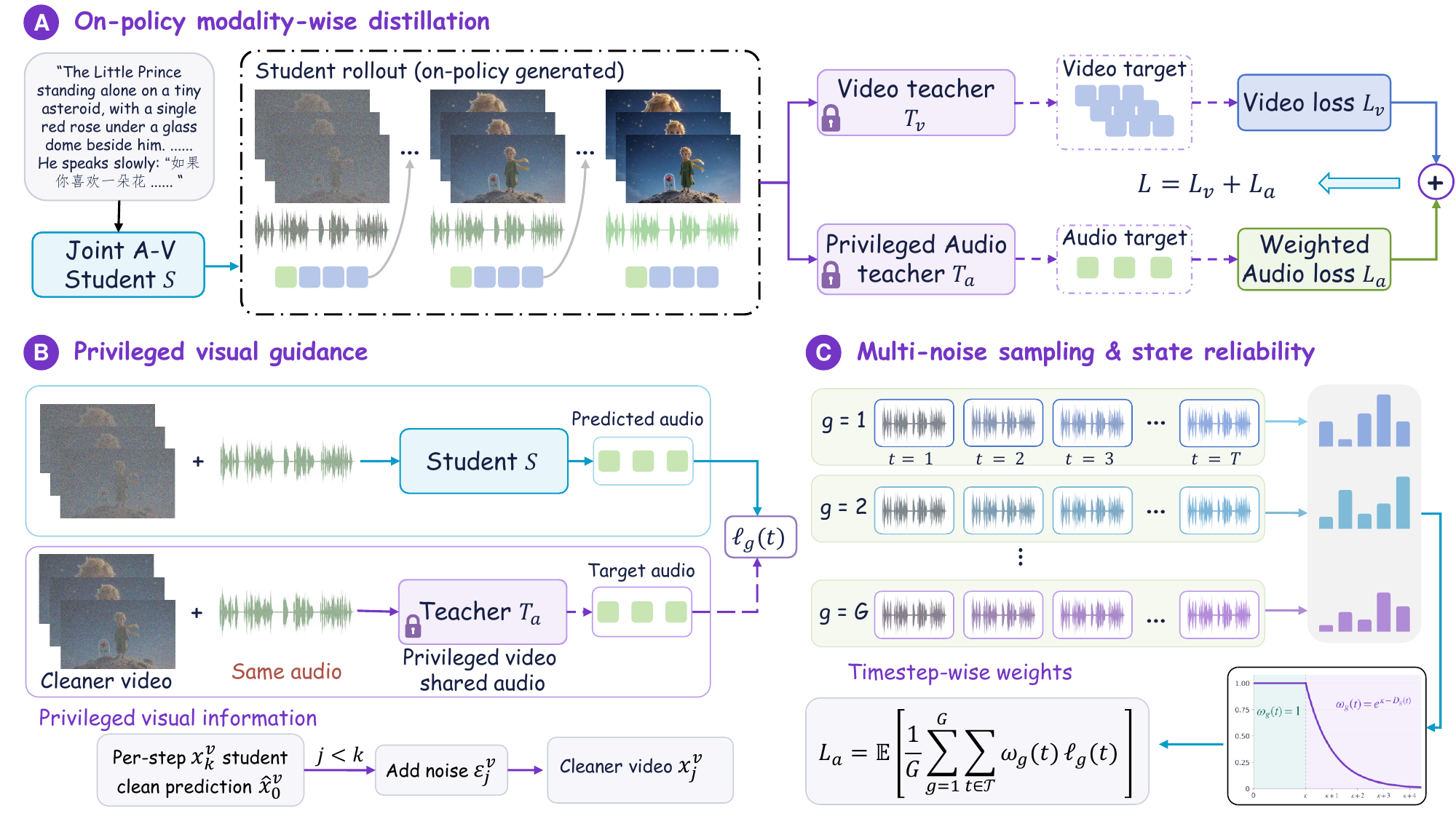}
    \caption{\textbf{Overview of OmniOPD.} (A) On-policy modality-wise distillation transfers knowledge from frozen video and audio teachers to a joint audio-video student using student-generated rollouts. (B) Privileged visual guidance provides the audio teacher with cleaner video context while sharing the same audio state with the student, yielding more informative audio supervision. (C) Multi-noise sampling expands audio supervision across noise realizations and denoising steps, while state-reliability weighting controls the contribution of unreliable learning signals.}
    \label{fig:pipeline}
\end{figure}

\section{OmniOPD}
\label{sec:method}
Motivated by the above observations and analysis in Sec.~\ref{sec:motivation}, we propose OmniOPD in Fig.~\ref{fig:pipeline}, a modality-wise OPD framework designed for balanced and stable audio-video training, which introduces asymmetric noise for privileged visual guidance in Sec.~\ref{sec:PVG}, grouped sampling to expand audio supervision coverage in Sec.~\ref{sec:multi_noise}, and importance weighting to stabilize audio learning in Sec.~\ref{sec:dtr}.

\subsection{Modality-Wise OPD with Privileged Visual Guidance}
\label{sec:PVG}

We adopt modality-wise on-policy distillation to integrate audio and video teachers into a unified student. Let $S$ denote the student and $T_a,T_v$ the frozen modality-specialized teachers. Along student-generated joint trajectories $\tau\sim P_\theta$, each teacher supervises its respective modality at the joint state $\mathbf{x}_{t}=(x^a_{t},x^v_{t})$ for each noise level $t\in\mathcal T$.
We formulate the distillation objective in $\mathbf{x}_0$-space to supervise each modality $m\in\{a,v\}$:
\begin{equation}
\mathcal L_m(\theta)
=
\mathbb E_{\tau\sim P_\theta}
\left[
\sum_{t\in\mathcal T}
\left\|
\hat{x}_{0}^{S}
-\hat{x}_{0}^{T_m}
\right\|_2^2
\right],
\end{equation}
where $\hat{x}_{0}^{S}$ and $\hat{x}_{0}^{T_m}$ are the student and teacher clean predictions at state $\mathbf{x}_{t}$. The relationship to transition divergence is derived in Appendix~\ref{app:x0}.

To strengthen audio supervision, we introduce privileged visual guidance based on our observation that informative video context supports audio learning. We choose a lower teacher-video noise level $q(t)$ satisfying $0<q(t)<t$. Using the rollout student's clean video prediction $\hat{x}^{S}_t$, we construct
\begin{equation}
\widetilde{x}^v_{q(t)}
=(1-q(t))z_v^t+q(t)\epsilon_v,
\qquad
\epsilon_v\sim\mathcal N(0,I).
\end{equation}
The student and audio teacher receive
\begin{equation}
\mathbf{x}^S_t=(x^a_t,x^v_t),
\qquad
\mathbf{x}^T_t=(x^a_t,\widetilde{x}^v_{q(t)}).
\end{equation}
Both models share the same audio state, while the teacher receives a cleaner video condition. This asymmetry provides privileged visual information that helps the teacher produce more informative audio targets.
Let $\widetilde{x}_0^{T_a}$ denote the audio target predicted by $T_a$ using privileged visual context. The audio objective is
\begin{equation}
\mathcal L_a^{\mathrm{PVG}}(\theta)
=
\mathbb E_{\tau\sim P_\theta}
\left[
\sum_{t\in\mathcal T}
\left\|
\hat{x}_0^{S}
-\widetilde{x}_0^{T_a}
\right\|_2^2
\right].
\end{equation}

\subsection{Multi-Noise Audio Distillation}
\label{sec:multi_noise}

A straightforward way to increase audio supervision is to sample more trajectories for each prompt. We independently sample $G$ joint audio-video trajectories $\tau_g\sim P_\theta(\cdot\mid c)$ for each prompt $c$. This directly increases the number of supervised audio tokens per prompt and broadens supervision coverage across diverse joint states. We apply privileged visual guidance within each trajectory to obtain the corresponding audio teacher targets.
For each sampled trajectory, we compute the audio distillation loss at each timestep:
\begin{equation}
\ell_g(t)
=
\left\|
\hat{x}_{0,g}^{S}
-\widetilde{x}_{0,g}^{T_a}
\right\|_2^2,
\end{equation}
where $\hat{x}_{0,g}^{S}$ and $\widetilde{x}_{0,g}^{T_a}$ denote the student prediction and teacher target of trajectory $g$. The grouped objective is
\begin{equation}
\mathcal L_a^{\mathrm{G}}(\theta)
=
\mathbb E
\left[
\frac{1}{G}
\sum_{g=1}^{G}
\sum_{t\in\mathcal T}
\ell_g(t)
\right],
\end{equation}
where the expectation is over the sampled trajectories and privileged visual conditions. Grouped sampling increases supervision diversity and broadens distribution coverage, reducing the sampling variance of the estimated audio objective.

\subsection{Timestep-Wise Importance Weighting}
\label{sec:dtr}

The reliability of audio supervision varies across trajectories and timesteps. We quantify the local teacher-student mismatch using the audio transition divergence:
\begin{equation}
D_g(t)
=
D_{\mathrm{KL}}(p_S(\cdot\mid \mathbf{x}_{t})\|p_{T_a}(\cdot\mid \mathbf{x}_{t}))
=
w(t)\left\|
v_{S}(\mathbf{x}_t^S;\theta)-v_{T_a}(\mathbf{x}_t^T)
\right\|_2^2,
\label{eq:audio_transition_kl}
\end{equation}
where $p_{S}$ and $p_{T_a}$ are the audio transition kernels under the student and teacher's respective visual conditions, with a common transition parameterization and shared covariance. Here, $v_{S}$ and $v_{T_a}$ denote their audio velocity predictions, and $w(t)$ is the schedule-dependent coefficient defined in Sec.~\ref{sec:opd}.
We assign each supervision term a importance weight:
\begin{equation}
\omega_g(t)
=
\operatorname{sg}\!\left[
\min\left(1,e^{\kappa-D_g(t)}\right)
\right],
\end{equation}
where $\kappa\geq0$ defines the trust-region boundary. Steps within this region retain full supervision, while those with larger transition mismatch are smoothly down-weighted.

\subsection{Overall Objective}
\label{sec:overall}

OmniOPD combines privileged visual guidance, grouped sampling, and timestep-wise importance weighting to strengthen audio supervision during joint distillation. Privileged visual context helps the audio teacher produce more informative targets, while sampling multiple joint trajectories for each prompt increases the number and diversity of supervised audio tokens. Importance weighting further adjusts the contribution of each supervision term according to its local teacher-student transition mismatch. The final objective combines the stabilized audio loss with video distillation:
\begin{equation}
\mathcal L_{OmniOPD}(\theta)
=
L_v(\theta) +
\mathbb E
\left[
\frac{1}{G}
\sum_{g=1}^{G}
\sum_{t\in\mathcal T}
\omega_g(t)\ell_g(t)
\right].
\end{equation}

\section{Experiments}
\label{sec:exp}

\begin{table*}[t]
\centering
\caption{Quantitative comparison on JavisBench. Higher is better for $\uparrow$ metrics, and lower is better for $\downarrow$ metrics. Best results are in \textbf{bold}, and second-best results are \underline{underlined}.}
\label{tab:javisbench}
\setlength{\tabcolsep}{5pt}
\renewcommand{\arraystretch}{1.1}
\resizebox{\textwidth}{!}{
\begin{tabular}{l*{10}{c}}
\toprule
& \multicolumn{2}{c}{AV-Quality}
& \multicolumn{4}{c}{Text-Consistency}
& \multicolumn{2}{c}{AV-Consistency}
& \multicolumn{2}{c}{AV-Synchrony} \\
\cmidrule(lr){2-3}
\cmidrule(lr){4-7}
\cmidrule(lr){8-9}
\cmidrule(lr){10-11}
Model
& VQ$\uparrow$
& AQ$\uparrow$
& TV-IB$\uparrow$
& TA-IB$\uparrow$
& CLIP$\uparrow$
& CLAP$\uparrow$
& AV-IB$\uparrow$
& AVHScore$\uparrow$
& JavisScore$\uparrow$
& DeSync$\downarrow$ \\
\midrule

\multicolumn{11}{l}{\textit{-- T2A+A2V}} \\
TempoTkn~\citep{yariv2024diverse}
& -- & -- & 0.084 & -- & 0.205 & --
& 0.139 & 0.122 & 0.103 & 1.532 \\
TPoS~\citep{jeong2023power}
& -- & -- & 0.201 & -- & 0.229 & --
& 0.124 & 0.129 & 0.095 & 1.493 \\
\midrule

\multicolumn{11}{l}{\textit{-- T2V+V2A}} \\
ReWaS~\citep{jeong2025read}
& -- & -- & -- & 0.123 & -- & 0.280
& 0.110 & 0.104 & 0.079 & 1.071 \\
See\&Hear~\citep{xing2024seeing} 
& -- & -- & -- & 0.129 & -- & 0.263
& 0.160 & 0.143 & 0.112 & 1.099 \\
FoleyCrafter~\citep{zhang2026foleycrafter}
& -- & -- & -- & 0.149 & -- & 0.383
& 0.193 & 0.186 & 0.151 & 0.952 \\
MMAudio~\citep{cheng2025mmaudio}
& -- & -- & -- & 0.160 & -- & 0.407
& 0.198 & 0.182 & 0.150 & 0.849 \\
\midrule

\multicolumn{11}{l}{\textit{-- T2AV}} \\
JavisDiT~\citep{liu2025javisdit}
& 1.291 & 4.478 & 0.263 & 0.143 & 0.302 & 0.391
& 0.197 & 0.179 & 0.154 & 1.039 \\
UniVerse-1~\citep{wang2025universe}
& 1.357 & 4.839 & 0.272 & 0.111 & 0.309 & 0.245
& 0.104 & 0.098 & 0.077 & 0.929 \\
JavisDiT++~\citep{liu2026javisditpp}
& 1.462 & 5.049 & \textbf{0.282} & 0.164 & \textbf{0.316} & 0.424
& 0.198 & 0.184 & 0.159 & 0.832 \\
LTX-2.3~\citep{hacohen2026ltx2}
& 2.714 & 5.252 & 0.279 & 0.178 & 0.308 & 0.429
& 0.256 & 0.251 & 0.222 & 0.584 \\
LTX-2.3+OmniNFT~\citep{zhang2026omninft}
& 3.216 & 5.564 & 0.274 & \underline{0.183} & 0.309 & 0.442
& 0.268 & 0.261 & 0.236 & 0.547 \\
\midrule
Video-Teacher
& 3.121 & 5.216 & 0.276 & 0.182 & 0.304 & 0.446
& \underline{0.275} & \underline{0.269} & \underline{0.238} & 0.465 \\
Audio-Teacher
& 2.929 & 5.576 & 0.276 & 0.184 & 0.307 & 0.470
& \textbf{0.282} & \textbf{0.279} & \textbf{0.251} & \textbf{0.448} \\
Ours OmniOPD
& \textbf{3.347} & \underline{5.449} & 0.275 & \textbf{0.192} & \underline{0.310} & \underline{0.447} & 0.262 & 0.258 & 0.232 & \underline{0.457} \\
\bottomrule
\end{tabular}
}
\end{table*}

\begin{figure}
    \centering
    \includegraphics[width=1.0\linewidth]{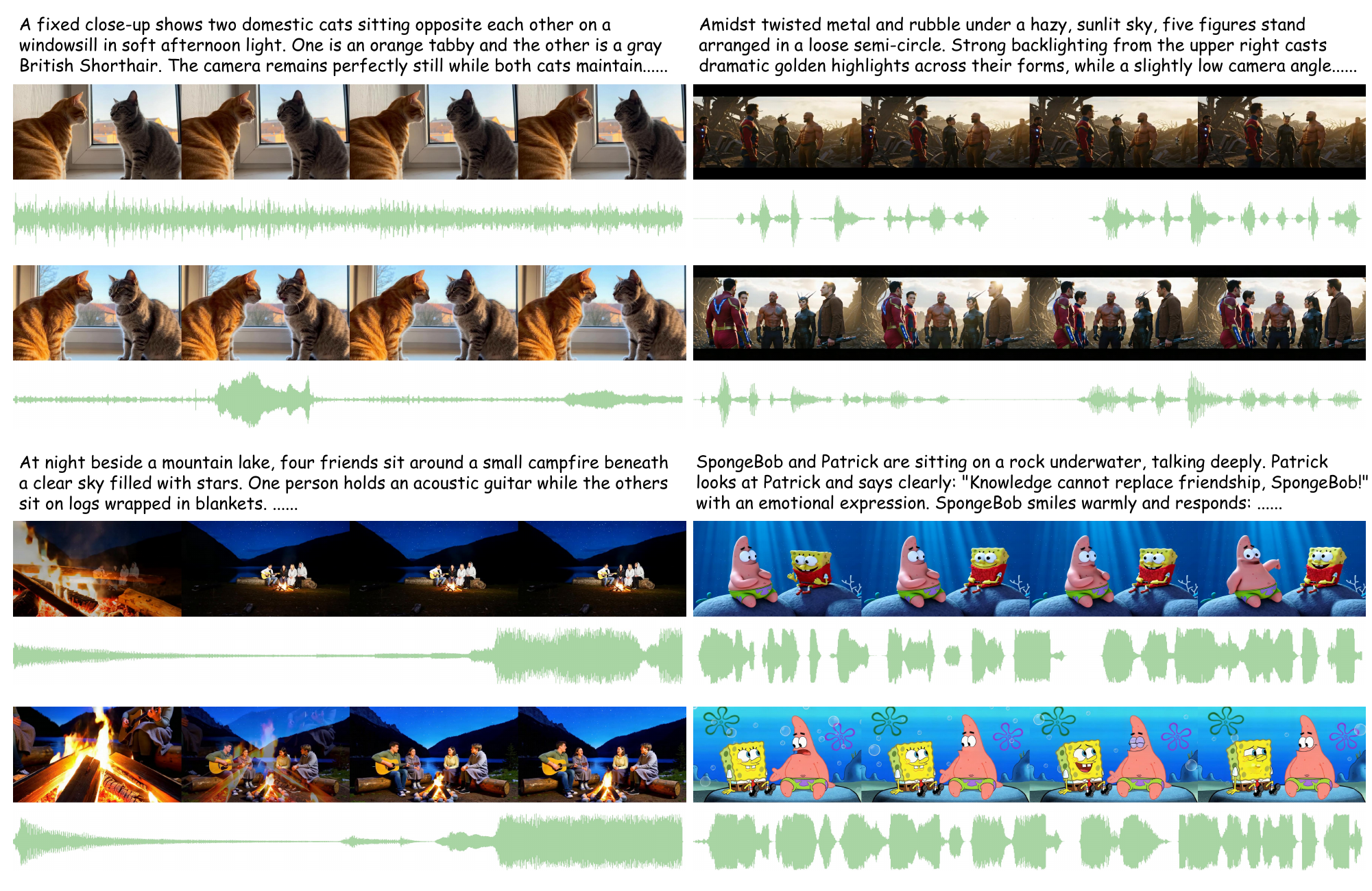}
    \caption{Qualitative examples of joint audio-video generation by OmniOPD. These examples demonstrate improvements in audio diversity, voice–character alignment in multi-character scenes, visual aesthetics, and audio fidelity.}
    \label{fig:mainfig}
\end{figure}

We evaluate OmniOPD on joint audio-video generation, focusing on whether modality-specialized teacher capabilities can be effectively integrated into a single student. We first describe the reward models, teacher configurations, and evaluation benchmarks, followed by the main results and ablation studies.

\subsection{Experimental Setup}
\label{sec:setup}

\paragraph{Reward models and base model.}
Following OmniNFT~\citep{zhang2026omninft}, we adopt five complementary rewards. HPSv3~\citep{ma2025hpsv3} and VideoAlign~\citep{liu2025improving} evaluate visual quality, while Audiobox Aesthetics~\citep{tjandra2025meta} evaluates audio quality. CLAP~\citep{wu2023large} measures audio-text alignment, and a reward derived from DeSync~\citep{iashin2024synchformer} measures audio-video synchronization. We use LTX-2 (22B)~\citep{hacohen2026ltx2}, a dual-stream joint audio-video DiT, as both the student initialization and the base model for teacher training.

\paragraph{Teacher training and RL variants.}
We independently train a video teacher and an audio teacher using DiffusionNFT~\citep{zheng2025diffusionnft} with LoRA~\citep{hu2022lora} fine-tuning. The video teacher is optimized with HPSv3, VideoAlign, and the synchronization reward, while the audio teacher is optimized with Audiobox Aesthetics, CLAP, and the synchronization reward. Both teachers remain frozen during distillation and supervise their respective modalities. To examine alternative strategies for improving both modalities, we additionally consider joint and sequential RL variants. \textbf{All reward} jointly optimizes all five rewards. All reward 1, All reward 2, and All reward 3 use weight ratios of $1{:}1{:}1{:}1{:}1$, $1{:}1{:}2{:}2{:}2$, and $1{:}1{:}4{:}4{:}4$, respectively, in the order HPSv3, VideoAlign, Audiobox, CLAP, and synchronization reward. These configurations progressively increase the weights of audio and synchronization rewards to examine whether greater emphasis on these objectives alleviates the imbalance in joint optimization. \textbf{Cascade} first optimizes video and then audio: both variants are initialized from the video teacher and subsequently fine-tuned using audio rewards. Cascade 1 updates both modality branches, whereas Cascade 2 updates only the audio-branch parameters while keeping the video branch frozen. This comparison examines how restricting second-stage updates affects audio improvement and the retention of previously learned video capabilities.

\begin{table*}[t]
\centering
\caption{Quantitative comparison on JavisBench under different experimental settings.}
\label{tab:javisbench1}
\setlength{\tabcolsep}{5pt}
\renewcommand{\arraystretch}{1.1}
\resizebox{\textwidth}{!}{
\begin{tabular}{l*{10}{c}}
\toprule
& \multicolumn{2}{c}{AV-Quality}
& \multicolumn{4}{c}{Text-Consistency}
& \multicolumn{2}{c}{AV-Consistency}
& \multicolumn{2}{c}{AV-Synchrony} \\
\cmidrule(lr){2-3}
\cmidrule(lr){4-7}
\cmidrule(lr){8-9}
\cmidrule(lr){10-11}
Model
& VQ$\uparrow$
& AQ$\uparrow$
& TV-IB$\uparrow$
& TA-IB$\uparrow$
& CLIP$\uparrow$
& CLAP$\uparrow$
& AV-IB$\uparrow$
& AVHScore$\uparrow$
& JavisScore$\uparrow$
& DeSync$\downarrow$ \\
\midrule

LTX-2.3
& 2.714 & 5.252 & \underline{0.279} & 0.178 & 0.308 & 0.429
& 0.256 & 0.251 & 0.222 & 0.584 \\
LTX-2.3+GDPO
& 2.713 & \textbf{5.577} & \textbf{0.280} & \underline{0.181} & \textbf{0.313} & 0.439
& 0.251 & 0.249 & 0.221 & 0.563 \\
Cascade
& 3.131 & 5.235 & 0.273 & 0.179 & 0.306 & \textbf{0.463} & \textbf{0.274} & \textbf{0.269} & \textbf{0.241} & \textbf{0.432} \\
Vanilla OPD
& \underline{3.174} & 5.342 & 0.275 & 0.180 & 0.309 & 0.433 & 0.256 & 0.250 & 0.226 & 0.583 \\
Ours OmniOPD
& \textbf{3.347} & \underline{5.449} & 0.275 & \textbf{0.192} & \underline{0.310} & \underline{0.447} & \underline{0.262} & \underline{0.258} & \underline{0.232} & \underline{0.457} \\
\bottomrule
\end{tabular}
}
\end{table*}
\paragraph{Evaluation benchmarks.}
We evaluate on JavisBench~\citep{liu2025javisdit} and VBench~\citep{huang2023vbench}. Following the evaluation organization of OmniNFT, we report four complementary aspects on JavisBench: audio-video quality, text consistency, audio-video semantic consistency, and temporal synchronization. The reported metrics include VQ and AQ; TV-IB, TA-IB, CLIP, and CLAP; AV-IB and AVHScore; and JavisScore and DeSync. Lower DeSync indicates better synchronization. VBench provides additional evaluation of video generation across dimensions such as imaging quality, aesthetic quality, subject consistency.

\subsection{Main Results}
\label{sec:main_results}

\begin{figure*}[t]
    \centering

    \sbox{\ablationbox}{%
        \small
        \setlength{\tabcolsep}{4pt}%
        \renewcommand{\arraystretch}{1.12}%
        \resizebox{0.60\textwidth}{!}{%
            \begin{tabular}{lccccc}
                \toprule
                Setting & VQ$\uparrow$ & AQ$\uparrow$
                & TV-IB$\uparrow$ & TA-IB$\uparrow$ & JavisScore$\uparrow$ \\
                \midrule
                \multicolumn{6}{l}{\textit{(a) Asymmetric-noise shift}} \\
                \midrule
                $s=0$   & 3.174 & 5.342 & 0.275 & 0.180 & 0.226 \\
                $s=1.5$ & 3.332 & 5.376 & 0.275 & 0.189 & 0.232 \\
                $s=2.0$ & 3.354 & 5.353 & 0.274 & 0.183 & 0.229 \\                         \midrule
                \multicolumn{6}{l}{\textit{(b) Number of rollout groups}} \\
                \midrule
                $G=1$ & 3.174 & 5.342 & 0.275 & 0.180 & 0.226 \\
                $G=2$ & 3.334 & 5.223 & 0.275 & 0.186 & 0.217 \\
                $G=3$ & 3.315 & 5.393 & 0.276 & 0.188 & 0.233 \\
                \midrule
                \multicolumn{6}{l}{\textit{(c) Trust-region soft ratio}} \\
                \midrule
                $p=0.75$       & 3.322 & 5.381 & 0.276 & 0.189 & 0.236\\
                $p=0.5$        & 3.321 & 5.407 & 0.273 & 0.184 & 0.235 \\
                w/o weighting  & 3.174 & 5.342 & 0.275 & 0.180 & 0.226 \\
                \bottomrule
            \end{tabular}%
        }%
    }
    \setlength{\ablationheight}{%
        \dimexpr\ht\ablationbox+\dp\ablationbox\relax
    }

    \begin{minipage}[t]{0.60\textwidth}
        \vspace{0pt}
        \centering
        \usebox{\ablationbox}\par
        \captionsetup{position=bottom,skip=6pt}
        \captionof{table}{Ablation studies on the asymmetric-noise shift,
        trust-region soft ratio, and number of rollout groups.}
        \label{tab:ablations}
    \end{minipage}
    \hfill
    \begin{minipage}[t]{0.37\textwidth}
        \vspace{0pt}
        \centering
        \includegraphics[height=\ablationheight]{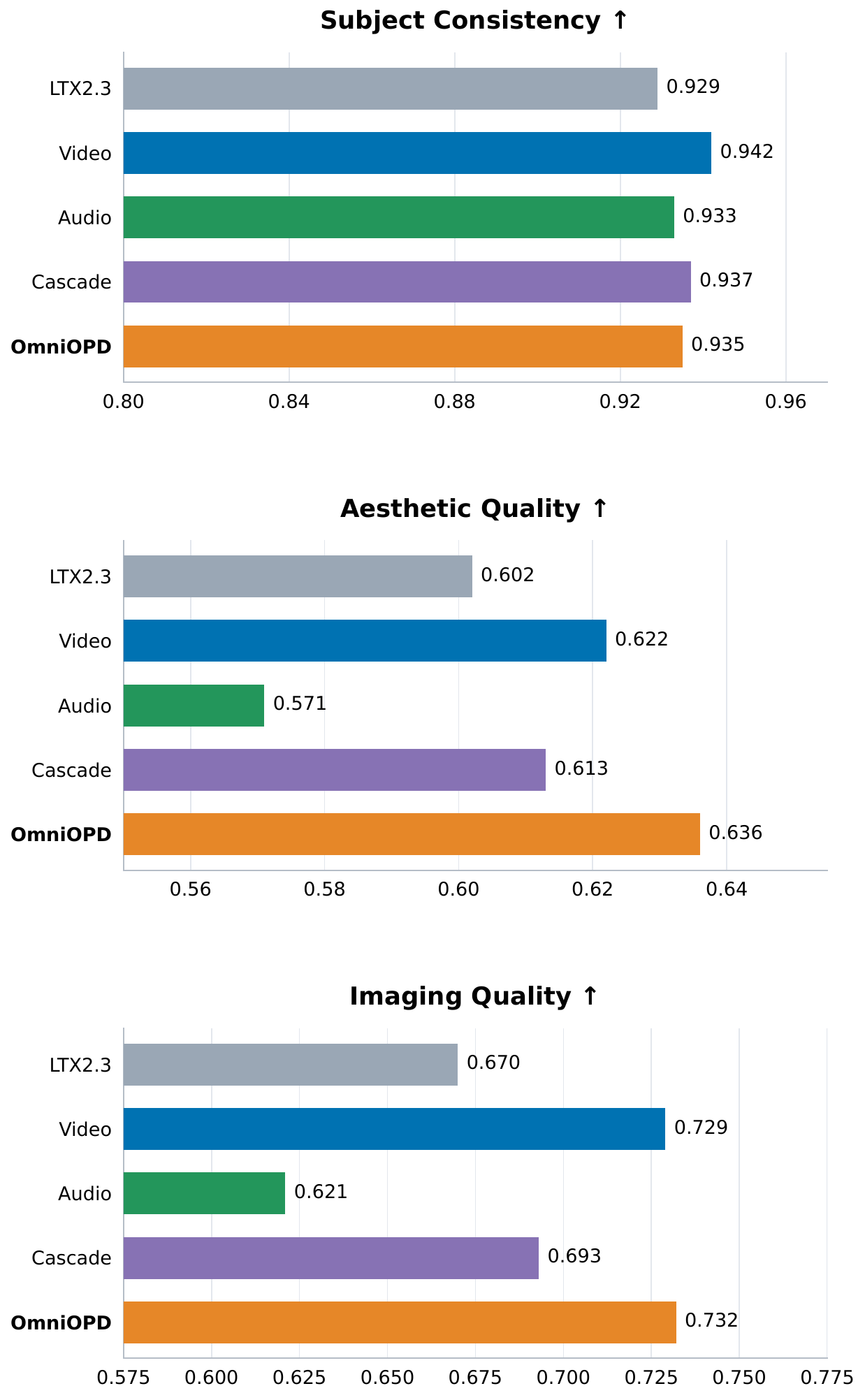}\par
        \captionsetup{position=bottom,skip=6pt}
        \captionof{figure}{Quantitative comparison on VBench.}
        \label{fig:vbench}
    \end{minipage}
\end{figure*}

\paragraph{Benchmark performance.}
Table~\ref{tab:javisbench} and Fig.~\ref{fig:vbench} report the results on JavisBench and VBench, respectively. As illustrated in Fig.~\ref{fig:mainfig}, OmniOPD generates high-fidelity, diverse audio, strengthening capabilities that are underdeveloped in the base model. Privileged visual guidance further improves voice-character alignment in multi-character scenes. OmniOPD achieves the best audio-video balance among the compared methods, substantially improving video quality while maintaining strong audio performance (Table~\ref{tab:javisbench1}). These results suggest that stabilizing audio learning also facilitates greater gains in video quality.

\subsection{Ablation Studies}
\label{sec:ablation}
\paragraph{Privileged visual guidance.}
We examine how the noise level of the video input to the teacher affects audio distillation. Alongside the matched-noise baseline $q(t)=t$, we evaluate timestep-shifting strategies that provide cleaner visual context. Specifically, the \textbf{shift} strategy selects the teacher-video timestep as
\begin{equation}
    q(t) \;=\; \frac{t}{\,t + e^{s}(1-t)\,},
    \qquad s>0,
    \label{eq:flowshift}
\end{equation}
where larger $s$ produces a lower teacher-video noise level. The results show that providing cleaner visual context substantially improves audio quality while also yielding gains in video quality in Table~\ref
{tab:ablations}(a), highlighting the benefits of visual guidance for joint optimization. 

\paragraph{Audio supervision coverage.}
We vary the number of branches $G$ used to aggregate audio supervision, with $G=1$ corresponding to a single branch. This comparison evaluates how additional supervision samples affect audio performance and training stability. As shown in Table~\ref{tab:ablations}(b), increasing audio supervision improves audio performance, while more stable audio learning also benefits video quality. Fig.~\ref{fig:mainfig} further illustrates that multi-group sampling broadens the diversity of supervision, helping strengthen capabilities that are underdeveloped in the base model.

\paragraph{Reliability weighting.}
We compare uniform audio supervision with the proposed Timestep-Wise Importance Weighting and vary the parameter controlling its weighting strength. As shown in Table~\ref{tab:ablations}(c), timestep-wise weighting of audio supervision stabilizes audio learning, helping mitigate the lasting degradation in video quality associated with unstable audio learning.

\section{Conclusion}

We presented OmniOPD, motivated by a comprehensive analysis of audio-video coupling that reveals cross-modal interference and sparse, stochastic audio supervision. OmniOPD combines asymmetric supervision, privileged visual guidance, and reliability-aware multi-branch learning for joint distillation. Experiments on JavisBench and VBench show improved video generation alongside strong audio performance, demonstrating more balanced audio-video distillation.

\bibliography{iclr2027_conference}

@misc{GRPO,
      title={DeepSeekMath: Pushing the Limits of Mathematical Reasoning in Open Language Models}, 
      author={Zhihong Shao and Peiyi Wang and Qihao Zhu and Runxin Xu and Junxiao Song and Xiao Bi and Haowei Zhang and Mingchuan Zhang and Y. K. Li and Y. Wu and Daya Guo},
      year={2024},
      eprint={2402.03300},
      archivePrefix={arXiv},
      primaryClass={cs.CL},
      url={https://arxiv.org/abs/2402.03300}, 
}

@inproceedings{GDPO,
title={{GDPO}: Group reward-Decoupled Normalization Policy Optimization for Multi-reward {RL} Optimization},
author={Shih-Yang Liu and Xin Dong and Ximing Lu and Shizhe Diao and Peter Belcak and Mingjie Liu and Min-Hung Chen and Hongxu Yin and Yu-Chiang Frank Wang and Kwang-Ting Cheng and Yejin Choi and Jan Kautz and Pavlo Molchanov},
booktitle={Forty-third International Conference on Machine Learning},
year={2026},
url={https://openreview.net/forum?id=Y1rHxA58If}
}

@article{flowopd,
  title={{Flow-OPD}: On-Policy Distillation for Flow Matching Models},
  author={Fang, Zhen and Huang, Wenxuan and Zeng, Yu and Zhao, Yiming and Chen, Shuang and Feng, Kaituo and Lin, Yunlong and Chen, Lin and Chen, Zehui and Cao, Shaosheng and Zhao, Feng},
  journal={arXiv preprint arXiv:2605.08063},
  year={2026},
  url={https://arxiv.org/abs/2605.08063}
}

@inproceedings{
chefer2026selfflow,
title={Self-Supervised Flow Matching for Scalable Multi-Modal Synthesis},
author={Hila Chefer and Patrick Esser and Dominik Lorenz and Dustin Podell and Vikash Raja and Vinh Tong and Antonio Torralba and Robin Rombach},
booktitle={Forty-third International Conference on Machine Learning},
year={2026},
url={https://openreview.net/forum?id=HoThWhfxiK}
}

@article{diffusionopd,
  title={{DiffusionOPD}: A Unified Perspective of On-Policy Distillation in Diffusion Models},
  author={Li, Quanhao and Yu, Junqiu and Jiang, Kaixun and Wei, Yujie and Xing, Zhen and Li, Pandeng and Chu, Ruihang and Zhang, Shiwei and Liu, Yu and Wu, Zuxuan},
  journal={arXiv preprint arXiv:2605.15055},
  year={2026},
  url={https://arxiv.org/abs/2605.15055}
}

@article{zhang2026omninft,
  title={{OmniNFT}: Modality-wise Omni Diffusion Reinforcement for Joint Audio-Video Generation},
  author={Zhang, Guohui and Ma, XiaoXiao and Huang, Jie and Xu, Hang and Yu, Hu and Fu, Siming and Li, Yuming and Xue, Zeyue and Song, Lin and Huang, Haoyang and Duan, Nan and Zhao, Feng},
  journal={arXiv preprint arXiv:2605.12480},
  year={2026},
  url={https://arxiv.org/abs/2605.12480}
}

@INPROCEEDINGS{ma2025hpsv3,
  author={Ma, Yuhang and Wu, Xiaoshi and Sun, Keqiang and Li, Hongsheng},
  booktitle={2025 IEEE/CVF International Conference on Computer Vision (ICCV)}, 
  title={HPSv3: Towards Wide-Spectrum Human Preference Score}, 
  year={2025},
  volume={},
  number={},
  pages={15086-15095},
  doi={10.1109/ICCV51701.2025.01400}}

@inproceedings{liu2025improving,
title={Improving Video Generation with Human Feedback},
author={Jie Liu and Gongye Liu and Jiajun Liang and Ziyang Yuan and Xiaokun Liu and Mingwu Zheng and Xiele Wu and Qiulin Wang and Menghan Xia and Xintao Wang and Xiaohong Liu and Fei Yang and Pengfei Wan and Di ZHANG and Kun Gai and Yujiu Yang and Wanli Ouyang},
booktitle={The Thirty-ninth Annual Conference on Neural Information Processing Systems},
year={2025},
url={https://openreview.net/forum?id=nHkg4yc7SP}
}

@INPROCEEDINGS{iashin2024synchformer,
  author={Iashin, Vladimir and Xie, Weidi and Rahtu, Esa and Zisserman, Andrew},
  booktitle={ICASSP 2024 - 2024 IEEE International Conference on Acoustics, Speech and Signal Processing (ICASSP)}, 
  title={Synchformer: Efficient Synchronization From Sparse Cues}, 
  year={2024},
  volume={},
  number={},
  pages={5325-5329},
  doi={10.1109/ICASSP48485.2024.10448489}}

@INPROCEEDINGS{tjandra2025meta,
  author={Tjandra, Andros and Wu, Yi-Chiao and Guo, Baishan and Hoffman, John and Ellis, Brian and Vyas, Apoorv and Shi, Bowen and Chen, Sanyuan and Le, Matt and Zacharov, Nick and Wood, Carleigh and Lee, Ann and Hsu, Wei-Ning},
  booktitle={2025 IEEE Automatic Speech Recognition and Understanding Workshop (ASRU)}, 
  title={Meta Audiobox Aesthetics: Unified Automatic Assessment for Speech, Music and Sound}, 
  year={2025},
  volume={},
  number={},
  pages={1-8},
  doi={10.1109/ASRU65441.2025.11434623}}

@INPROCEEDINGS{wu2023large,
  author={Wu, Yusong and Chen, Ke and Zhang, Tianyu and Hui, Yuchen and Berg-Kirkpatrick, Taylor and Dubnov, Shlomo},
  booktitle={ICASSP 2023 - 2023 IEEE International Conference on Acoustics, Speech and Signal Processing (ICASSP)}, 
  title={Large-Scale Contrastive Language-Audio Pretraining with Feature Fusion and Keyword-to-Caption Augmentation}, 
  year={2023},
  volume={},
  number={},
  pages={1-5},
  doi={10.1109/ICASSP49357.2023.10095969}}

@inproceedings{zheng2025diffusionnft,
 author = {Zheng, Kaiwen and Chen, Huayu and Ye, Haotian and Wang, Haoxiang and Zhang, Qinsheng and Jiang, Kai and Su, Hang and Ermon, Stefano and Zhu, Jun and Liu, Ming-Yu},
 booktitle = {International Conference on Learning Representations},
 editor = {C. Vondrick and B. Hariharan and C. Raffel and L. Pinto and D. Yang and A. Faust},
 pages = {134129--134150},
 title = {DiffusionNFT: Online Diffusion Reinforcement with Forward Process},
 url = {https://proceedings.iclr.cc/paper_files/paper/2026/file/d8e68ddfe22520b45b8fb8d5cbde5a21-Paper-Conference.pdf},
 volume = {2026},
 year = {2026}
}

@inproceedings{
liu2025javisdit,
title={JavisDiT: Joint Audio-Video Diffusion Transformer with Hierarchical Spatio-Temporal Prior Synchronization},
author={Kai Liu and Wei Li and Lai Chen and Shengqiong Wu and Yanhao Zheng and Jiayi Ji and Fan Zhou and Jiebo Luo and Ziwei Liu and Hao Fei and Tat-Seng Chua},
booktitle={The Fourteenth International Conference on Learning Representations},
year={2026},
url={https://openreview.net/forum?id=y7HV7KT3Bd}
}

@InProceedings{ruan2022mmdiffusion,
    author    = {Ruan, Ludan and Ma, Yiyang and Yang, Huan and He, Huiguo and Liu, Bei and Fu, Jianlong and Yuan, Nicholas Jing and Jin, Qin and Guo, Baining},
    title     = {MM-Diffusion: Learning Multi-Modal Diffusion Models for Joint Audio and Video Generation},
    booktitle = {Proceedings of the IEEE/CVF Conference on Computer Vision and Pattern Recognition (CVPR)},
    month     = {June},
    year      = {2023},
    pages     = {10219-10228}
}

@article{wang2025universe,
  title={UniVerse-1: Unified Audio-Video Generation via Stitching of Experts},
  author={Wang, Duomin and Zuo, Wei and Li, Aojie and Chen, Ling-Hao and Liao, Xinyao and Zhou, Deyu and Yin, Zixin and Dai, Xili and Jiang, Daxin and Yu, Gang},
  journal={arXiv preprint arXiv:2509.06155},
  year={2025}
}

@inproceedings{black2023ddpo,
 author = {Black, Kevin and Janner, Michael and Du, Yilun and Kostrikov, Ilya and Levine, Sergey},
 booktitle = {International Conference on Learning Representations},
 editor = {B. Kim and Y. Yue and S. Chaudhuri and K. Fragkiadaki and M. Khan and Y. Sun},
 pages = {4965--4987},
 title = {Training Diffusion Models with Reinforcement Learning},
 url = {https://proceedings.iclr.cc/paper_files/paper/2024/file/14f75513f0f1ca01de1e826b52e6b840-Paper-Conference.pdf},
 volume = {2024},
 year = {2024}
}

@inproceedings{fan2023dpok,
 author = {Fan, Ying and Watkins, Olivia and Du, Yuqing and Liu, Hao and Ryu, Moonkyung and Boutilier, Craig and Abbeel, Pieter and Ghavamzadeh, Mohammad and Lee, Kangwook and Lee, Kimin},
 booktitle = {Advances in Neural Information Processing Systems},
 doi = {10.52202/075280-3497},
 editor = {A. Oh and T. Naumann and A. Globerson and K. Saenko and M. Hardt and S. Levine},
 pages = {79858--79885},
 publisher = {Curran Associates, Inc.},
 title = {DPOK: Reinforcement Learning for Fine-tuning Text-to-Image Diffusion Models},
 url = {https://proceedings.neurips.cc/paper_files/paper/2023/file/fc65fab891d83433bd3c8d966edde311-Paper-Conference.pdf},
 volume = {36},
 year = {2023}
}

@inproceedings{liu2025flowgrpo,
 author = {Liu, Jie and Liu, Gongye and Liang, Jiajun and Li, Yangguang and Liu, Jiaheng and Wang, Xintao and Wan, Pengfei and ZHANG, Di and Ouyang, Wanli},
 booktitle = {Advances in Neural Information Processing Systems},
 doi = {10.52202/085713-1362},
 editor = {D. Belgrave and C. Zhang and H. Lin and R. Pascanu and P. Koniusz and M. Ghassemi and N. Chen},
 pages = {40783--40818},
 publisher = {Curran Associates, Inc.},
 title = {Flow-GRPO: Training Flow Matching Models via Online RL},
 url = {https://proceedings.neurips.cc/paper_files/paper/2025/file/3a10c46572628d58cb44fb705f25cbbf-Paper-Conference.pdf},
 volume = {38, Main Conference},
 year = {2025}
}

@article{xue2025dancegrpo,
  title={DanceGRPO: Unleashing GRPO on Visual Generation},
  author={Xue, Zeyue and Wu, Jie and Gao, Yu and Kong, Fangyuan and Zhu, Lingting and Chen, Mengzhao and Liu, Zhiheng and Liu, Wei and Guo, Qiushan and Huang, Weilin and others},
  journal={arXiv preprint arXiv:2505.07818},
  year={2025}
}

@inproceedings{agarwal2023gkd,
 author = {Agarwal, Rishabh and Vieillard, Nino and Zhou, Yongchao and Stanczyk, Piotr and Ramos Garea, Sabela and Geist, Matthieu and Bachem, Olivier},
 booktitle = {International Conference on Learning Representations},
 editor = {B. Kim and Y. Yue and S. Chaudhuri and K. Fragkiadaki and M. Khan and Y. Sun},
 pages = {21246--21263},
 title = {On-Policy Distillation of Language Models: Learning from Self-Generated Mistakes},
 url = {https://proceedings.iclr.cc/paper_files/paper/2024/file/5be69a584901a26c521c2b51e40a4c20-Paper-Conference.pdf},
 volume = {2024},
 year = {2024}
}

@inproceedings{
gu2023minillm,
title={Mini{LLM}: Knowledge Distillation of Large Language Models},
author={Yuxian Gu and Li Dong and Furu Wei and Minlie Huang},
booktitle={The Twelfth International Conference on Learning Representations},
year={2024},
url={https://openreview.net/forum?id=5h0qf7IBZZ}
}

@InProceedings{ko2024distillm,
  title = 	 {{D}isti{LLM}: Towards Streamlined Distillation for Large Language Models},
  author =       {Ko, Jongwoo and Kim, Sungnyun and Chen, Tianyi and Yun, Se-Young},
  booktitle = 	 {Proceedings of the 41st International Conference on Machine Learning},
  pages = 	 {24872--24895},
  year = 	 {2024},
  editor = 	 {Salakhutdinov, Ruslan and Kolter, Zico and Heller, Katherine and Weller, Adrian and Oliver, Nuria and Scarlett, Jonathan and Berkenkamp, Felix},
  volume = 	 {235},
  series = 	 {Proceedings of Machine Learning Research},
  month = 	 {21--27 Jul},
  publisher =    {PMLR},
  url = 	 {https://proceedings.mlr.press/v235/ko24c.html},
}

@misc{ma2026mopd,
      title={MOPD: Multi-Teacher On-Policy Distillation for Capability Integration in LLM Post-Training}, 
      author={Wenhan Ma and Jianyu Wei and Liang Zhao and Hailin Zhang and Bangjun Xiao and Lei Li and Qibin Yang and Bofei Gao and Yudong Wang and Rang Li and Jinhao Dong and Zhifang Sui and Fuli Luo},
      year={2026},
      eprint={2606.30406},
      archivePrefix={arXiv},
      primaryClass={cs.CL},
      url={https://arxiv.org/abs/2606.30406}, 
}

@misc{gao2026openmopd,
      title={Open-MOPD: Diagnosing and Fixing Capability Imbalance in Multi-Teacher On-Policy Distillation}, 
      author={Huan-ang Gao and Haohan Chi and Yong Yan and Shiyuan Feng and Hanlin Wu and Zheng Jiang and Bingxiang He and Wei-Ying Ma and Ya-Qin Zhang and Hao Zhou},
      year={2026},
      eprint={2608.19098},
      archivePrefix={arXiv},
      primaryClass={cs.LG},
      url={https://arxiv.org/abs/2608.19098}, 
}

@misc{xing2026tropd,
      title={Trust Region On-Policy Distillation}, 
      author={Xingrun Xing and Haoqing Wang and Boyan Gao and Ziheng Li and Yehui Tang},
      year={2026},
      eprint={2606.01249},
      archivePrefix={arXiv},
      primaryClass={cs.LG},
      url={https://arxiv.org/abs/2606.01249}, 
}

@misc{wei2026stepopd,
      title={STEP-OPD: Rethinking Output Targets and Internal Dynamics in On-Policy Distillation for Diffusion Models}, 
      author={Qingyan Wei and Guangzhao Li and Xiaobing Tu and Yinggui Wang and Xiantao Zhang and Jinkui Ren and Xiaohong Liu and Linfeng Zhang},
      year={2026},
      eprint={2608.04887},
      archivePrefix={arXiv},
      primaryClass={cs.CV},
      url={https://arxiv.org/abs/2608.04887}, 
}

@misc{zhou2026diffusionopsd,
      title={On-Policy Self-Distillation in Diffusion Models}, 
      author={Wei Zhou and Xiongwei Zhu and Lingdong Kong and Bo Chen and Lei Zhang and Yongyuan Liang and Xiaoxia Hou and Ye Tian and Xian Sun and Yingshuo Wang and Linfeng Li and Shengqiong Wu and Leigang Qu and Feng Li and Wei Liu and Julian McAuley and Tat-Seng Chua},
      year={2026},
      eprint={2608.24646},
      archivePrefix={arXiv},
      primaryClass={cs.CV},
      url={https://arxiv.org/abs/2608.24646}, 
}

@inproceedings{
lipman2023flow,
title={Flow Matching for Generative Modeling},
author={Yaron Lipman and Ricky T. Q. Chen and Heli Ben-Hamu and Maximilian Nickel and Matthew Le},
booktitle={The Eleventh International Conference on Learning Representations },
year={2023},
url={https://openreview.net/forum?id=PqvMRDCJT9t}
}

@article{thinkingmachines2026opd,
  author = {Kevin Lu and Thinking Machines Lab},
  title = {On-Policy Distillation},
  journal = {Thinking Machines Lab: Connectionism},
  year = {2025},
  note = {https://thinkingmachines.ai/blog/on-policy-distillation},
  doi = {10.64434/tml.20251026},
}

@misc{hacohen2026ltx2,
      title={LTX-2: Efficient Joint Audio-Visual Foundation Model}, 
      author={Yoav HaCohen and Benny Brazowski and Nisan Chiprut and Yaki Bitterman and Andrew Kvochko and Avishai Berkowitz and Daniel Shalem and Daphna Lifschitz and Dudu Moshe and Eitan Porat and Eitan Richardson and Guy Shiran and Itay Chachy and Jonathan Chetboun and Michael Finkelson and Michael Kupchick and Nir Zabari and Nitzan Guetta and Noa Kotler and Ofir Bibi and Ori Gordon and Poriya Panet and Roi Benita and Shahar Armon and Victor Kulikov and Yaron Inger and Yonatan Shiftan and Zeev Melumian and Zeev Farbman},
      year={2026},
      eprint={2601.03233},
      archivePrefix={arXiv},
      primaryClass={cs.CV},
      url={https://arxiv.org/abs/2601.03233}, 
}

@misc{low2025ovi,
      title={Ovi: Twin Backbone Cross-Modal Fusion for Audio-Video Generation}, 
      author={Chetwin Low and Weimin Wang and Calder Katyal},
      year={2025},
      eprint={2510.01284},
      archivePrefix={arXiv},
      primaryClass={cs.MM},
      url={https://arxiv.org/abs/2510.01284}, 
}

@misc{liu2026javisditpp,
      title={JavisDiT++: Unified Modeling and Optimization for Joint Audio-Video Generation}, 
      author={Kai Liu and Yanhao Zheng and Kai Wang and Shengqiong Wu and Rongjunchen Zhang and Jiebo Luo and Dimitrios Hatzinakos and Ziwei Liu and Hao Fei and Tat-Seng Chua},
      year={2026},
      eprint={2602.19163},
      archivePrefix={arXiv},
      primaryClass={cs.CV},
      url={https://arxiv.org/abs/2602.19163}, 
}

@inproceedings{xu2023imagereward,
  title={ImageReward: learning and evaluating human preferences for text-to-image generation},
  author={Xu, Jiazheng and Liu, Xiao and Wu, Yuchen and Tong, Yuxuan and Li, Qinkai and Ding, Ming and Tang, Jie and Dong, Yuxiao},
  booktitle={Proceedings of the 37th International Conference on Neural Information Processing Systems},
  pages={15903--15935},
  year={2023}
}

@InProceedings{wallace2023diffusion,
    author    = {Wallace, Bram and Dang, Meihua and Rafailov, Rafael and Zhou, Linqi and Lou, Aaron and Purushwalkam, Senthil and Ermon, Stefano and Xiong, Caiming and Joty, Shafiq and Naik, Nikhil},
    title     = {Diffusion Model Alignment Using Direct Preference Optimization},
    booktitle = {Proceedings of the IEEE/CVF Conference on Computer Vision and Pattern Recognition (CVPR)},
    month     = {June},
    year      = {2024},
    pages     = {8228-8238}
}

@inproceedings{
hu2022lora,
title={Lo{RA}: Low-Rank Adaptation of Large Language Models},
author={Edward J Hu and yelong shen and Phillip Wallis and Zeyuan Allen-Zhu and Yuanzhi Li and Shean Wang and Lu Wang and Weizhu Chen},
booktitle={International Conference on Learning Representations},
year={2022},
url={https://openreview.net/forum?id=nZeVKeeFYf9}
}

@INPROCEEDINGS{huang2023vbench,
  author={Huang, Ziqi and He, Yinan and Yu, Jiashuo and Zhang, Fan and Si, Chenyang and Jiang, Yuming and Zhang, Yuanhan and Wu, Tianxing and Jin, Qingyang and Chanpaisit, Nattapol and Wang, Yaohui and Chen, Xinyuan and Wang, Limin and Lin, Dahua and Qiao, Yu and Liu, Ziwei},
  booktitle={2024 IEEE/CVF Conference on Computer Vision and Pattern Recognition (CVPR)}, 
  title={VBench: Comprehensive Benchmark Suite for Video Generative Models}, 
  year={2024},
  volume={},
  number={},
  pages={21807-21818},
  doi={10.1109/CVPR52733.2024.02060}}

@inproceedings{yariv2024diverse,
  title={Diverse and aligned audio-to-video generation via text-to-video model adaptation},
  author={Yariv, Guy and Gat, Itai and Benaim, Sagie and Wolf, Lior and Schwartz, Idan and Adi, Yossi},
  booktitle={Proceedings of the AAAI Conference on Artificial Intelligence},
  volume={38},
  pages={6639--6647},
  year={2024}
}

@inproceedings{jeong2023power,
  title={The power of sound (tpos): Audio reactive video generation with stable diffusion},
  author={Jeong, Yujin and Ryoo, Wonjeong and Lee, Seunghyun and Seo, Dabin and Byeon, Wonmin and Kim, Sangpil and Kim, Jinkyu},
  booktitle={Proceedings of the IEEE/CVF international conference on computer vision},
  pages={7822--7832},
  year={2023}
}

@inproceedings{jeong2025read,
  title={Read, watch and scream! sound generation from text and video},
  author={Jeong, Yujin and Kim, Yunji and Chun, Sanghyuk and Lee, Jiyoung},
  booktitle={Proceedings of the AAAI Conference on Artificial Intelligence},
  volume={39},
  pages={17590--17598},
  year={2025}
}

@inproceedings{xing2024seeing,
  title={Seeing and hearing: Open-domain visual-audio generation with diffusion latent aligners},
  author={Xing, Yazhou and He, Yingqing and Tian, Zeyue and Wang, Xintao and Chen, Qifeng},
  booktitle={Proceedings of the IEEE/CVF Conference on Computer Vision and Pattern Recognition},
  pages={7151--7161},
  year={2024}
}

@article{zhang2026foleycrafter,
  title={Foleycrafter: Bring silent videos to life with lifelike and synchronized sounds},
  author={Zhang, Yiming and Gu, Yicheng and Zeng, Yanhong and Xing, Zhening and Wang, Yuancheng and Wu, Zhizheng and Liu, Bin and Chen, Kai},
  journal={International Journal of Computer Vision},
  volume={134},
  number={1},
  pages={46},
  year={2026},
  publisher={Springer}
}

@inproceedings{cheng2025mmaudio,
  title={Mmaudio: Taming multimodal joint training for high-quality video-to-audio synthesis},
  author={Cheng, Ho Kei and Ishii, Masato and Hayakawa, Akio and Shibuya, Takashi and Schwing, Alexander and Mitsufuji, Yuki},
  booktitle={Proceedings of the Computer Vision and Pattern Recognition Conference},
  pages={28901--28911},
  year={2025}
}
\bibliographystyle{iclr2027_conference}

\appendix
\newpage
\section{Appendix}

\subsection{From Transition Divergence to Clean-Prediction Regression}
\label{app:x0}

We derive the relationship between transition divergence and
clean-prediction regression, omitting modality indices for clarity.
At a shared noisy state $x_{t}$ and noise level $t>0$, the clean
prediction of model $M\in\{S,T\}$ is
\begin{equation}
\hat{x}_M=x_{t}-t v_M,
\end{equation}
where $v_M$ denotes its velocity prediction. Since the student and
teacher share the same noisy state, subtraction gives
\begin{equation}
\hat{x}_S-\hat{x}_T=-t(v_S-v_T),
\qquad
\|v_S-v_T\|_2^2
=\frac{1}{t^2}\|\hat{x}_S-\hat{x}_T\|_2^2.
\end{equation}
Substituting this identity into Eq.~\eqref{eq:flow_local_kl} yields
\begin{equation}
D_{\mathrm{KL}}
=\frac{w(t)}{t^2}\|\hat{x}_S-\hat{x}_T\|_2^2
=\bar{w}(t)\|\hat{x}_S-\hat{x}_T\|_2^2,
\qquad
\bar{w}(t)=\frac{w(t)}{t^2}.
\label{eq:app_clean_kl}
\end{equation}

Thus, transition divergence corresponds to clean-prediction regression
with a schedule-dependent coefficient. We omit $\bar{w}(t)$ in the
regression objective, which changes the relative weighting across
timesteps and therefore defines a distillation surrogate rather than
an identical KL objective. This conversion also holds under different
visual conditions, provided that the student and teacher share the
same audio state and velocity-to-clean conversion coefficient.

\subsection{Limitations}
Although our work provides a comprehensive analysis of audio-video coupling and introduces effective strategies for joint optimization through OmniOPD, two related limitations remain. First, the teachers operate on student-generated states during on-policy distillation, which may differ from the input distributions encountered during teacher training. This input distribution shift can reduce the reliability of teacher predictions. Second, when student-generated states deviate substantially from the teachers' supported distributions, teacher predictions may become unreliable or misleading, rendering the distillation supervision ineffective. 

\begin{table}[t]
    \centering
    \caption{Prompts used for evaluation in Fig.~\ref{fig:mainfig}.}
    \label{tab:prompts}
    \begin{tabularx}{\linewidth}{cX}
        \toprule
        No. & Prompt \\
        \midrule
        1 & "A fixed close-up shows two domestic cats sitting opposite each other on a windowsill in soft afternoon light. One is an orange tabby and the other is a gray British Shorthair. The camera remains perfectly still while both cats maintain stable seated postures. The orange cat looks toward the gray cat and produces a gentle meow. After a brief pause, the gray cat slowly blinks and answers with a shorter, lower-pitched meow. Only subtle head, ear, and whisker movements occur. The audio contains the two clearly separated cat vocalizations with almost no other environmental sound." \\
        \midrule
        2 & "Amidst twisted metal and rubble under a hazy, sunlit sky, five figures stand arranged in a loose semi-circle. Strong backlighting from the upper right casts dramatic golden highlights across their forms, while a slightly low camera angle looks up at the group, emphasizing their presence. On the far left, a man in a red-and-gold suit with glowing blue accents stands next to a younger man in a red-and-blue suit. To their right, a muscular, bald man with reddish-brown skin and tattoos faces a woman in black leather armor with antennae-like headgear, while on the far right, a man in a brown jacket holds a futuristic weapon at his side.  The man in the red-and-gold suit turns his body slightly to look directly at the resolute-looking, muscular man opposite him and says quietly, “I can’t shake the feeling that this isn’t over yet.” In response, a faint smile appears on the muscular man’s face, and he replies in a calm, steady voice, “As long as we’re still standing together, we can get through anything.” The other three figures remain standing, their gazes shifting subtly as they listen in silence to the exchange.  The audio shows a quiet background ambience featuring the sound of soft wind moving through the ruins, punctuated by the faint, intermittent clinking of metal debris. The dialogue is clear, with the first man speaking quietly and the second replying in a calm, steady tone." \\
        \midrule
        3 & "At night beside a mountain lake, four friends sit around a small campfire beneath a clear sky filled with stars. One person holds an acoustic guitar while the others sit on logs wrapped in blankets. The camera begins with a close-up of burning wood and glowing embers before slowly pulling back to reveal the group and the dark lake behind them. The guitarist begins playing a gentle chord progression, and after several measures, a woman quietly starts singing. The others gradually join her during the chorus. The audio features detailed fire crackles in the foreground, occasional pieces of wood collapsing into the flames, and a naturally recorded acoustic guitar. The first singing voice is clear and close, while additional voices gradually create a small group harmony. Faint insects and distant water sounds remain audible behind the music." \\
        \midrule
        4 & "SpongeBob and Patrick are sitting on a rock underwater, talking deeply. Patrick looks at Patrick and says clearly: "Knowledge cannot replace friendship, SpongeBob!" with an emotional expression. SpongeBob smiles warmly and responds: "I agree, but I also really want that Krabby Patty!" The scene features professional lip-syncing and facial muscle movements.  Cinematic 2D animation, high-detail CGI. Soft underwater lighting, floating sea particles, 8k resolution, photorealistic textures for sponge and starfish skin." \\

        \bottomrule
    \end{tabularx}
\end{table}

\subsection{More Qualitative Examples}
\begin{figure}
    \centering
    \includegraphics[width=1.0\linewidth]{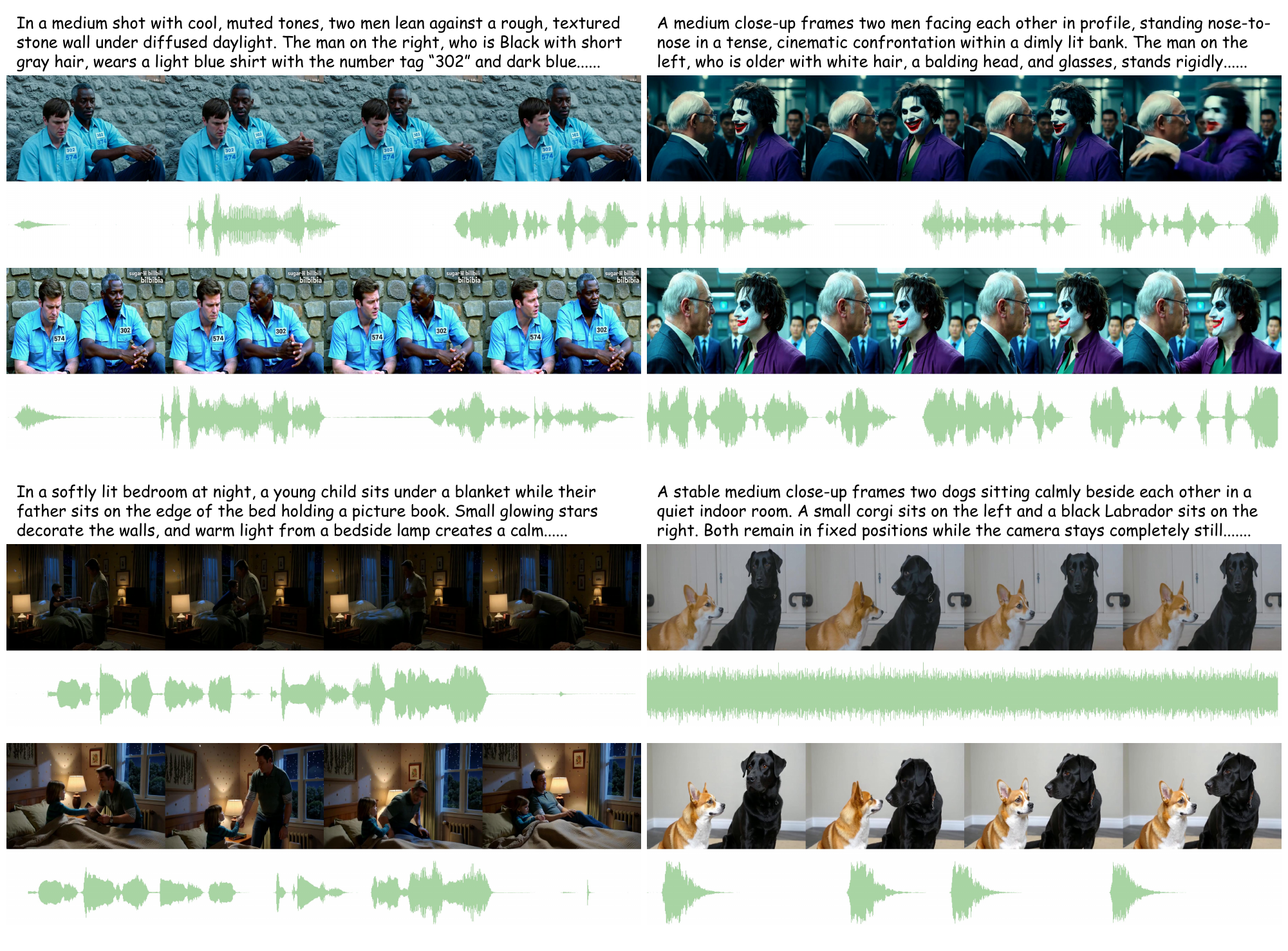}
    \caption{Qualitative examples of joint audio-video generation by OmniOPD. These examples demonstrate improvements in audio diversity, voice–character alignment in multi-character scenes, visual aesthetics, and audio fidelity.}
    \label{fig:mainfig2}
\end{figure}

\begin{figure}
    \centering
    \includegraphics[width=1.0\linewidth]{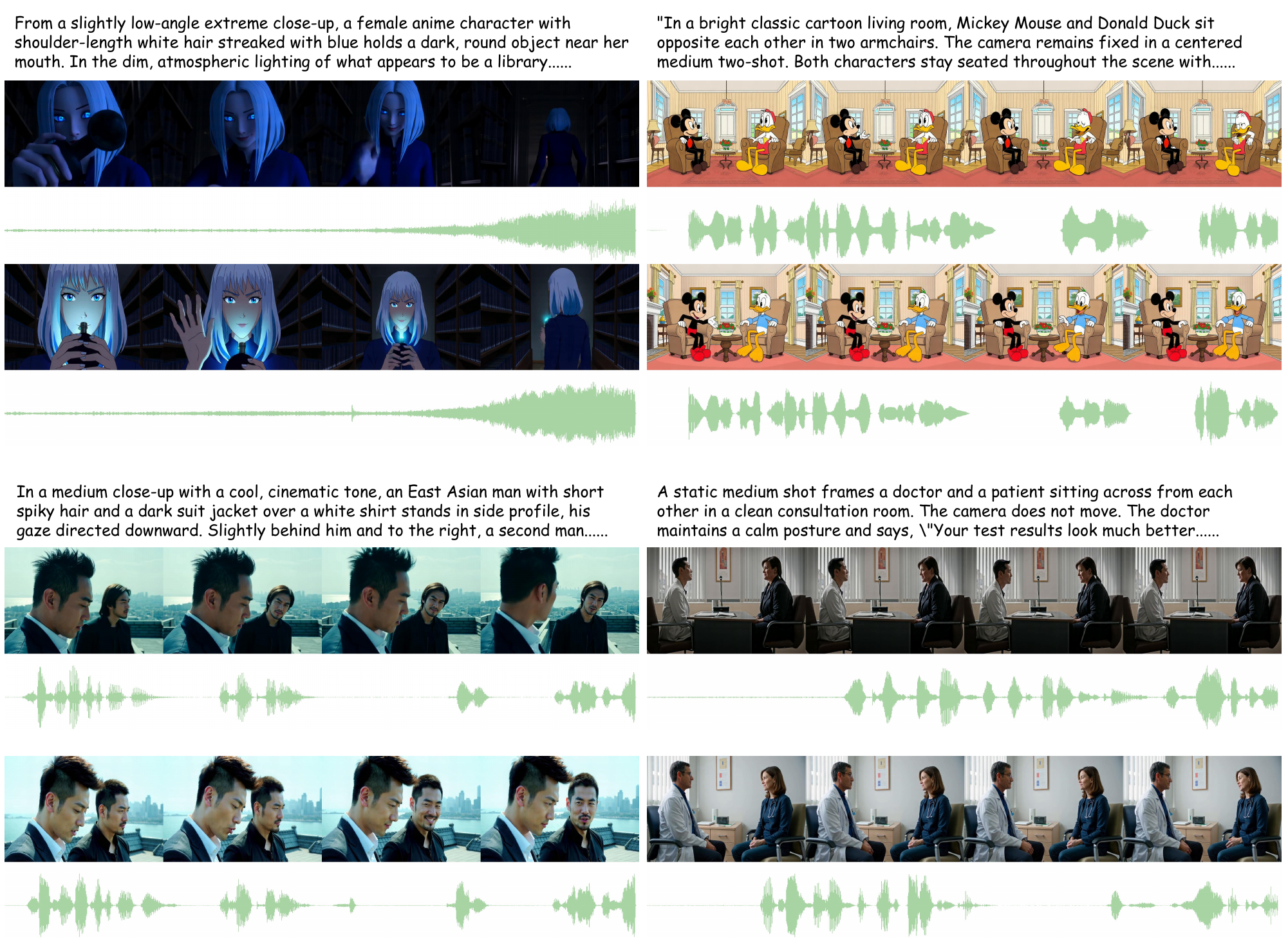}
    \caption{Qualitative examples of joint audio-video generation by OmniOPD. These examples demonstrate improvements in audio diversity, voice–character alignment in multi-character scenes, visual aesthetics, and audio fidelity.}
    \label{fig:mainfig3}
\end{figure}

Additional qualitative examples of joint audio-video generation by OmniOPD in Fig.~\ref{fig:mainfig2} and Fig.~\ref{fig:mainfig3}. These examples demonstrate improvements in audio diversity, voice-character alignment, visual aesthetics, and audio fidelity. Specific improvements include more legible text on characters' clothing, correct speaker attribution and lip synchronization in two-person conversations, improved audio generation for long-tail scenarios, and greater visual consistency without abrupt changes or disappearing elements.

\end{document}